\documentclass[]{article}

\title{MTHE 493 Thesis\\ Region Tracking in an Image Sequence\\ Preventing Driver Inattention}
\date{April 7, 2017}
\author{Matthew Kowal \\ Gillian Sandison \\ Len Yabuki-Soh \\ Raner la Bastide \\ \\Supervisor: Dr. Abdol-Reza Mansouri \\
Queen's University}

\usepackage{bbm}
\usepackage{amsmath}
\usepackage{amssymb}
\usepackage[margin = 1in]{geometry}
\usepackage[T1]{fontenc}
\usepackage[utf8x]{inputenc}
\usepackage{graphicx}
\usepackage{float} 
\usepackage{amsmath}
\usepackage{amssymb}
\usepackage{hyperref}
\usepackage{caption} 
\usepackage{siunitx}
\usepackage{cite}
\usepackage{url}
\usepackage{subfig}
\usepackage{mdframed}

\begin{document}
\pagenumbering{gobble}
\maketitle
\newpage
\pagenumbering{roman}
\section*{Abstract}
Driver inattention is a large problem on the roads around the world. The objective of this project was to develop an eye tracking algorithm with sufficient computational efficiency and accuracy, to successfully realize when the driver was looking away from the road for an extended period. The method of tracking involved the minimization of a functional, using the gradient descent and level set methods. The algorithm was then discretized and implemented using C and MATLAB. Multiple synthetic images, grey-scale and color images were tested using the final design, with a desired region coverage of 82\%. Further work is needed to decrease the computation time, increase the robustness of the algorithm, develop a small device capable of running the algorithm, as well as physically implement this device into various vehicles.

\newpage
\section*{Acknowledgments}
We would like to thank Professor Mansouri for all his guidance throughout the course of this project. His teachings and advice were invaluable to the success of the design.
\newpage
\tableofcontents
\newpage
\pagenumbering{arabic}
\section{Introduction}
%Impact of engineering/explains context of the system (Rubric Section): Identifies the context of the system in the Introduction by demonstrating a superior understanding of the societal, enterprise, and/or technical context of the system.
Safe driving requires the full attention of the driver, but distracted driving is very common despite this. In 2006, 78\% of traffic accidents were related to driver inattention \cite{prevent2011}. This alarming statistic is the motivation behind the overall goal of the project: to design a system that will monitor driver's eyes to ensure they are paying attention to the road, alerting them when it notices unacceptable eye positions. More specifically, the scope of this project is to design a system that will track a human eye in a sequence of images, and can be incorporated into a larger system that will reduce the negative effects of driver inattention on the road.
\subsection{Background}
Both region tracking and eye tracking have been approached in various ways. Related works and methods of approaching each of these problems were considered prior to commencing the design process. 
\subsubsection{Automated Region Tracking}
Region tracking is an important problem in computer vision that can be applied to numerous applications, from video surveillance to sports analytics. It can be very complex due to: the loss of information caused by projection of the 3D world on a 2D image; noise in images; complex object shapes; scene illumination changes; and real-time processing requirements.\\
\\Every method of region tracking requires a way of detecting the region of interest in every frame. One method that has been long used in tracking problems is point detection. These algorithms are concerned with tracking feature points \cite{Zheng}, corners, or edge segments \cite{Harris} of a given region. These methods use local detection of areas with high contrast. This approach is desirable because these points of interest are invariant to changes in illumination. However, this method requires numerous assumptions about the regions of interest, since only certain points or sections are actually being tracked. An alternate method of tracking the desired region is background subtraction \cite{yilmaz}. In this method, region tracking is achieved by building a representation of the scene called the background model, and then finding deviations from the model in the subsequent frames. This method requires significant changes in the position of the region, and also requires a completely static background, something which cannot be guaranteed in eye tracking since the background is a human subject.\\
\\Another approach to automated region tracking is motion-based tracking \cite{histreg}. This method requires motion information, and works well when the object undergoes minimal deformation, however when there are significant changes in the region's structure the efficiency in tracking the region decreases \cite{histreg}.
\subsubsection{Eye Tracking}
\begin{figure}[H]
\centering
\includegraphics[scale=0.8]{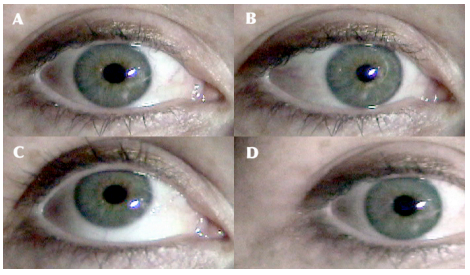}
\caption{The white dot on the right side of the pupil is the corneal reflection (A). Relative position of pupil and corneal reflection changes when the eye moves around its vertical and horizontal axes (B and C). The relative position does not change when the head moves but the eye is stable (D) \cite{richardson}.}
\label{fig:eyes}
\end{figure}

The movement of a human eye has been studied for years, with each method having unique trade-offs and difficulties. Early collections of data for a moving eye were obtained from observing the subject’s eye using a mirror, telescope or peep hole. As one might expect, these methods were inaccurate because of significant human error. The first notable advance in eye tracking came with the implementation of mechanical devices that were able to convert the eye movements into permanent objective records of motion \cite{richardson}. One technique following this advance used television cameras to scan the eye, where certain distinct features were electronically detected and localized \cite{richardson}. Since this method was particularly sensitive to contrast, another approach was taken which consists of scanning the boundary between the white sclera and the coloured iris, also known as the limbus \cite{richardson}. An alternate existing method is to scan for the lack of reflection from the eye. This approach presents difficulties when tracking vertical movement of the eye, since a large part of the iris is often covered by the eyelid  \cite{richardson}.\\
\\The aforementioned methods all record movement of the eye in relation to the head, which does not necessarily describe where the subject is looking. In an attempt to determine the subject's point of gaze, devices were used to restrain the subject's head with a chin rest or bite bar. Figure \ref{fig:eyes} shows the method of eye tracking not relative to the head.
Though more accurate gaze tracking was produced, applying this technique in a natural setting presents further difficulty. A specific example of such a device is the dual Purkinje image eye tracker. When the device applies light to the eye it produces four images. The dual Purkinje image eye tracker measures the difference between the first and fourth images by adjusting a series of mirrors with servomotors. The adjustments are made until the two images are superimposed on electronic photo-receptors. The dual Purkinje eye tracker is fast and accurate since it is only limited by the speed of the servomotors. The continuous analogue signal can be sampled at a rate of 300 Hz \cite{richardson}.
\\
\\One final method of eye tracking uses headband-eye trackers. This allows the subject to perform natural head and body movements, while still being able to obtain precise record of an observer’s point-of-regard \cite{richardson}. However, this comes with the obvious trade-off of having to wear such a device.

\subsection{Design Method}
The method of tracking an eye employed in this project is to design a suitable functional whose minimizer is the region of the eye consisting of the iris and the pupil. This approach is advantageous because it does not require any assumptions about the region based on feature points or contrasted edges of the boundary. It also does not require a completely static background. Additionally, the method does not use any motion information, but a model of the region being tracked. Finally, this method does not require any restraints on the subject whose eye is being tracked, making it practical for use in a vehicle while the subject is driving.  The objective is then to design a computer vision algorithm that will track the human eye from one image to the next.

\section{Design Approach}
\begin{figure}[H]
\centering
\includegraphics[scale=0.4]{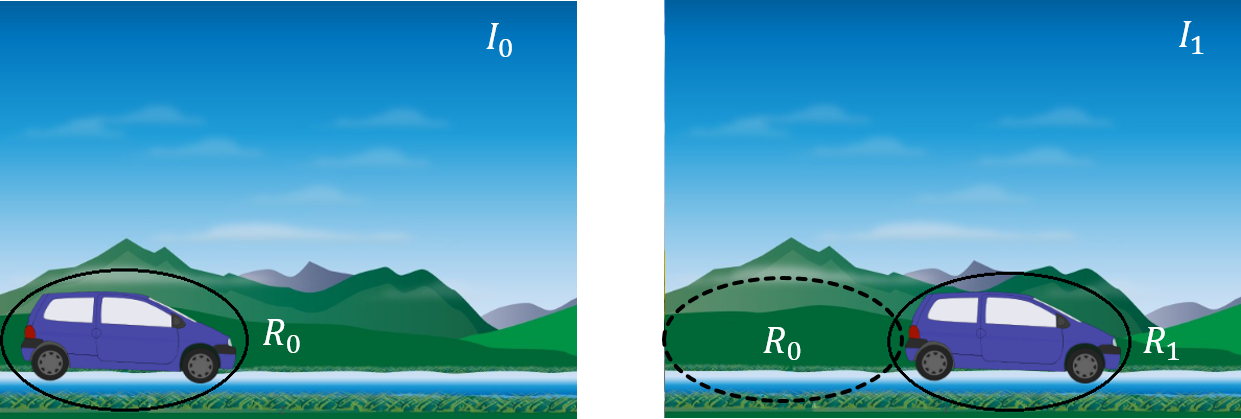}
\caption{Example sequence of images.}
\label{fig:probdef}
\end{figure}

%Problem Analysis/Identify Problem (Rubric Section): Demonstrates a superior understanding of all facets of the problem.
%Problem Analysis/Select Model (Rubric Section): Devises superior and/or novel models and methods of analysis.
%Design/Identify design problem (Rubric Section): Shows a superior understanding of the design problem and all constraints involved, including health and safety risks, applicable standards, economic, environmental, cultural and societal considerations.
%Design/Create process (Rubric Section): Is rigorous in creating correct processes for solving problems, understanding all approximations and assumptions
Let $\Omega\subset\mathbb{R}^2$ be the image domain and consider a sequence of images $I_0$, $I_1$,
\begin{align*}
I_0, I_1: \Omega \mapsto \mathbb{R}
\end{align*}
The region that will be tracked (the iris and pupil), $R_0\subset\Omega$, is defined in the initial image, $I_0$. The problem is then to find $R_1\subset\Omega$ in $I_1$ which corresponds to $R_0$. This problem is represented visually in Figure \ref{fig:probdef}.
This tracking problem is viewed as a minimization problem by defining a functional 
\begin{align*}
E:P(\Omega)\mapsto\mathbb{R}
\end{align*}
such that
\begin{align*}
R_1=arg\min_{P(\Omega)} E
\end{align*}
It is assumed that the regions of interest do not contain any holes and have smooth boundaries. Additionally it is assumed that $E$ has a minimizer, but note that this is not guaranteed. Then the functionals, $E$, can be defined as
\begin{align*}
E:\mathcal{C}\mapsto\mathbb{R}
\end{align*}
where
\begin{align*}
\mathcal{C}=\{\gamma:[0,1] \mapsto \mathbb{R}^2 | \gamma(0)=\gamma(1), \hat{\gamma} \in C^\infty \}
\end{align*}
where $\hat{\gamma}$ is the periodic extension of $\gamma$ to $\mathbb{R}$. The problem is then to design a suitable functional whose minimizer is the boundary of the region of interest. This variational approach was first introduced to computer vision by D. Mumford and J. Shah in 1989 \cite{Mumford}.

\subsection{Calculus of Variations and Gradient Descent}
Minimization of functionals follows the method illustrated by L. Gelfand and S. Fomin \cite{CalcVar}. Functionals considered for this method are of the form
\begin{align*}
E[\gamma]=\int_0^1L(s,x(s),y(s),\dot{x}(s),\dot{y}(s))ds
\end{align*}
where $\gamma(s)=(x(s),y(s))$. Minimization of $E$ leads to the gradient descent partial differential equation:
\begin{align*}
\frac{\partial \gamma}{\partial t}=-\frac{\delta E}{\delta \gamma}
\end{align*}
where
\begin{align*}
\frac{\delta E}{\delta \gamma}=
\begin{pmatrix}
\frac{\partial E}{\partial x}\\[6pt]
\frac{\partial E}{\partial y}
\end{pmatrix}
\end{align*}
and $\frac{\partial E}{\partial x}$, $\frac{\partial E}{\partial y}$ are computed by the Euler-Lagrange equations, seen below.
\begin{align*}
\frac{\partial E}{\partial x}=\bigg(\frac{\partial L}{\partial x}-\frac{d}{dt}\Big(\frac{\partial L}{\partial \dot{x}}\Big)\bigg)\\
\frac{\partial E}{\partial y}=\bigg(\frac{\partial L}{\partial y}-\frac{d}{dt}\Big(\frac{\partial L}{\partial \dot{y}}\Big)\bigg)
\end{align*}
Note that $t \mapsto E(\gamma(\cdot,t))$ is monotonically decreasing, hence $\gamma(t)$ is always approaching a minimizing curve.

\subsection{Level Set Representation}
\begin{figure}[H]
\centering
\includegraphics[scale=0.8]{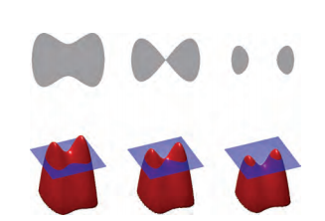}
\caption{Level set representation \cite{Colding}.}
\label{fig:levset}
\end{figure}

The curve can be viewed using a level set representation as described by J.A. Sethian \cite{Sethian}. Without loss of generality, $\frac{\delta E}{\delta \gamma}$ is taken to be $F\vec{N}$ for some $F$ where $\vec{N}$ is the unit normal. This is possible because any tangential component of $\frac{\delta E}{\delta \gamma}$ only results in reparameterization of $\gamma$. For $u:\Omega\mapsto\mathbb{R}$ with $u(\gamma)=0$
\begin{align*}
\frac{\partial \gamma}{\partial t}=F\vec{N}\Rightarrow\frac{\partial u}{\partial t}=F||\vec{\nabla}u|| 
\end{align*}
This level set implementation is illustrated in Figure \ref{fig:levset}, with the graph of $u$ shown in red. The  zero level set gives an implicit representation of the curve \cite{Colding}.
It is advantageous to represent the solution in this way because it provides numerical stability and topology independence \cite{Mansouri}.

\subsection{Discretization}
To allow for implementation of the continuous model, discretization of the system is performed according to J.A. Sethian \cite{Sethian}. This discretization is explained in the following example. Consider the one-dimensional wave equation $u_t(x,t)+u_x(x,t)=0$ with $u(x,0)=f(x)$.

\begin{figure}[H]
\centering
\includegraphics[scale=0.3]{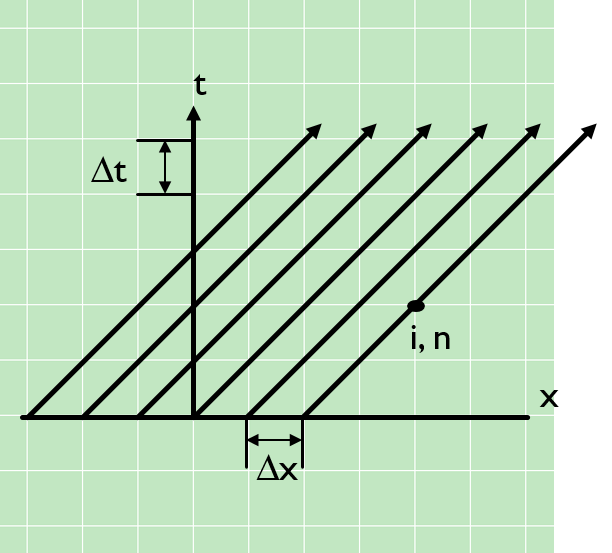}
\caption{Solution propagation of $u_t(x,t)+u_x(x,t)=0$.}
\label{fig:wave1}
\end{figure}

The solution to this equation is $u(x,t)=f(x-t)$, and is constant along the lines of slope $1$ drawn in the $x-t$ plane. In order to approximate the solution, begin by following the grid points in Figure \ref{fig:wave1}. The solution $u$ at time $t+\Delta t$ can be expanded as a Taylor series around to point $(x,t)$ and rearranged to give the forward difference operator for the time variable:
\begin{align*}
u_t=\frac{u(x,t+\Delta t)-u(x,t)}{\Delta t}+O(\Delta t)
\end{align*}
Defining
\begin{align*}
D^{+t}u=\frac{u(x,t+\Delta t)-u(x,t)}{\Delta t}
\end{align*}
gives
\begin{align*}
u_t=D^{+t}u+O(\Delta t)
\end{align*}
Similarly for the spatial derivative, define:
\begin{align*}
D^{+x}u=\frac{u(x+\Delta x,t)-u(x,t)}{\delta x}\\
D^{-x}u=\frac{u(x,t)-u(x-\Delta x,t)}{\delta x}\\
D^{0x}u=\frac{u(x+\Delta x,t)-u(x-\Delta x,t)}{2\Delta x}
\end{align*}
and then construct forward, backward, and centered Taylor series expansions in $x$ for the value $u$ around the point $(x,t)$:
\begin{align*}
u_{+x}=D^{+x}u+O(\Delta x)\\
u_{-x}=D^{-x}u+O(\Delta x)\\
u_{0x}=D^{0x}u+O(\Delta x)
\end{align*}
This leads to three schemes for computing the solution to the equation at time $n\Delta t$ and point $i\Delta x$. 
\begin{center}
Forward scheme: ${u_i}^{n+1}={u_i}^n-\Delta tD^{+x}{u_i}^n$\\
Backward scheme: ${u_i}^{n+1}={u_i}^n-\Delta tD^{-x}{u_i}^n$\\
Centered scheme: ${u_i}^{n+1}={u_i}^n-\Delta tD^{0x}{u_i}^n$
\end{center}

\begin{figure}[H]
\centering
\includegraphics[scale=0.3]{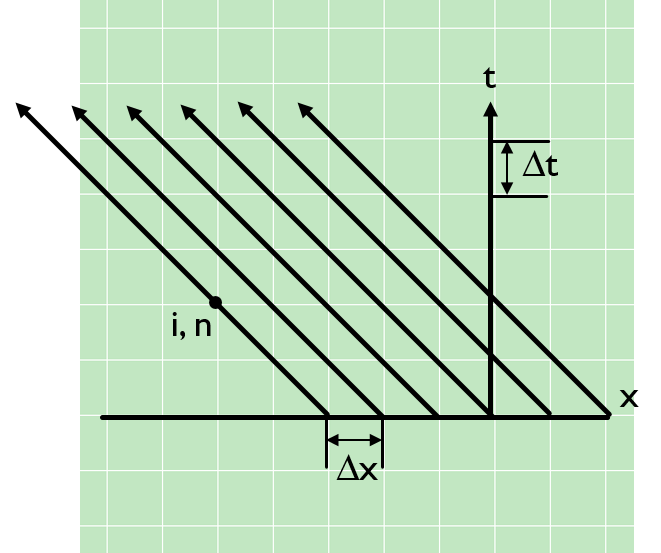}
\caption{Solution propagation of $u_t(x,t)-u_x(x,t)=0$.}
\label{fig:wave2}
\end{figure}

In this example, the backward scheme would be the one used because it is consistent with the theoretical solution, where information propagates from left to right. For the equation $u_t(x,t)-u_x(x,t)=0$ illustrated in Figure \ref{fig:wave2}, the solution is constant along the lines of slope $-1$, and hence the forward scheme should be used because it is consistent with the theoretical solution where information propagates from right to left.
The implemented discretization switches between the appropriate schemes (forward and backward), based on the theoretical solution, as follows:
\begin{align*}
\Delta^+=\big[max(D_{ij}^{-x},0)^2+min(D_{ij}^{+x},0)^2+max(D_{ij}^{-y},0)^2+min(D_{ij}^{+y},0)^2\big] ^\frac{1}{2}\\
\Delta^-=\big[max(D_{ij}^{+x},0)^2+min(D_{ij}^{-x},0)^2+max(D_{ij}^{+y},0)^2+min(D_{ij}^{-y},0)^2\big] ^\frac{1}{2}
\end{align*}
\subsection{Computer Programming Tools}
The algorithms designed using the afformentioned approach are implemented in the C programming language which interfaces with MATLAB. This is done because MATLAB has image processing tools that allow for easy conversion from image file to matrix and vice versa. MATLAB also allows for easy conversion between greyscale and colour images. C is used for the bulk of the computiation because it is more efficent than MATLAB for large computational loops. The interfacing between the two tools allows for the implementation to benefit from the strengths of both.
\section{Validation of Design Approach}
\begin{figure}[H]
\centering
\includegraphics[scale=1]{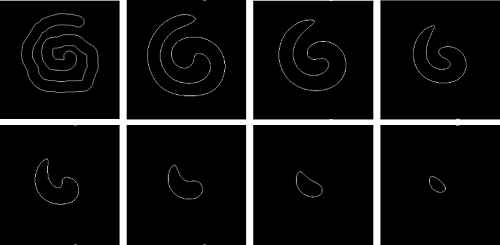}
\caption{Test implementation of design approach on the length functional.}
\label{fig:length}
\end{figure}

At approximately the half way point of the project, the design approach was implemented on the basic length functional $E=\int ||\dot{\gamma}(s)||ds$. This was done in order to validate the design approach: calculus of variations and gradient descent, level set transformation, and discritization. One set of results from the implementation is displayed in Figure \ref{fig:length}.
As expected, the length of the spiral curve was minimized by pushing out concavities and pushing in convexities. Since the above behaved as expected, the project was able to progress to designing functionals suitable for tracking a human eye. 
\section{Design Implementation}
\subsection{Designing Functionals}
\begin{figure}[H]
  \centering
  \subfloat[$I_0$]{\includegraphics[width=0.5\textwidth]{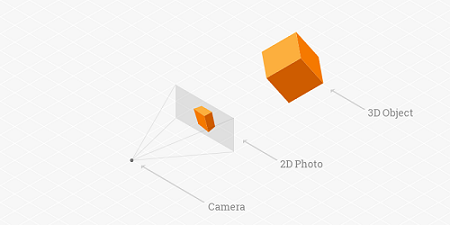}
  \label{fig:invar1}}
  \hfill
  \subfloat[$I_1$]{\includegraphics[width=0.5\textwidth]{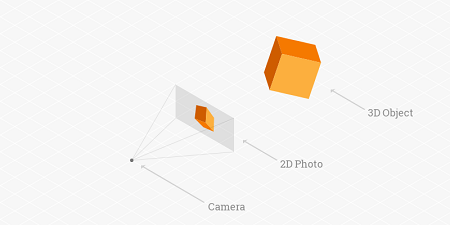}
  \label{fig:invar2}}
  \caption{Visualizing projections through the rotation of a 3D cube \cite{projections}.}
  \label{fig:invariance}
\end{figure}

When light travels from an object to a camera, it creates a projection of the 3D object on an image plane. The object always remains the same, what changes is how the object is represented in the image plane from one projection to the next. This concept is illustrated by two projections of a rotating cube in Figure \ref{fig:invariance}.
The problem is to track how the physical object is represented in the image plane from one image to the next, hence for this method of region tracking, functionals are designed based on the invariant characteristics of the region.
\subsection{Design Process}
\begin{figure}[H]
\centering
\includegraphics[scale=1]{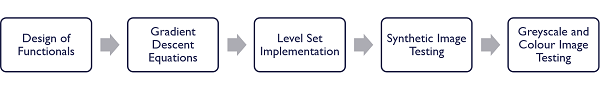}
\caption{Design process.}
\label{fig:designimp}
\end{figure}

\begin{figure}[H]
\centering
\includegraphics[scale=0.3]{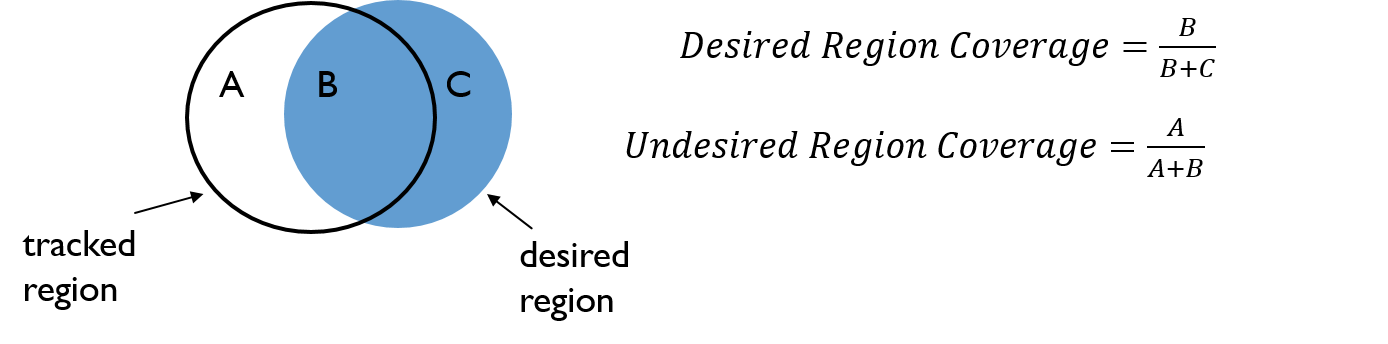}
\caption{Accuracy evaluation metrics.}
\label{fig:accurate}
\end{figure}

The design method is implemented following the steps outlined in Figure \ref{fig:designimp}. 
Before applying the design to greyscale and colour images of actual eyes, it was applied to synthetic images. This was done in order to ensure the design performed as it was intended, and that all the math was done correctly.\\
\\The performance of each designed functional was then assessed based primarily on its accuracy. The accuracy was quantified by the following two metrics: \textit{Desired Region Coverage}, and \textit{Undesired Region Coverage}. A diagram for better understanding these metrics can be seen in Figure \ref{fig:accurate}.
The \textit{Desired Region Coverage} measures in percentage the amount of the desired region that is covered by the tracked region. The \textit{Undesired Region Coverage} measures in percentage the amount of the tracked region that does not overlap the desired region. Clearly, the \textit{Desired Region Coverage} and \textit{Undesired Region Coverage} percentages should be maximized and minimized, respectively.\\
\\The results of each design were evaluated qualitatively, in order to determine why the above metrics were either too high or too low. Based on these evaluations, improvements were made to each design, and the design implementation process outlined above was performed iteratively until results suitable for the driver alertness application were achieved.

\section{Results}
%Design/Creates Simulations (Rubric Section): Creates tests superior and/or novel simulations, models, and/or prototypes at various points in design with complexity appropriate to design stage. 
%Design/Design Improvements (Rubric Sections): Is unusually assiduous in improving and reviewing designs to evaluate performance of the overall process i.e. demonstrates a recursive, and iterative design methodology that leads to more complex and/or applicable results. 
\subsection{Curve Regularization}
Every considered functional includes the following length term. 
\begin{align*}
 \lambda \int ||\dot{\gamma}(s)||  \,ds
\end{align*}
The purpose of this term is to regularize $\gamma$ by penalizing length. This is necessary because for any minimizing $\gamma$, there is a sequence of curves with irregular boundaries having arbitrary length that approach $\gamma$. By penalizing length, the functional is minimized at the most regular curve that contains the region of interest.
\subsection{Design \#1}
\begin{figure}[H]
\centering
\includegraphics[scale=0.8]{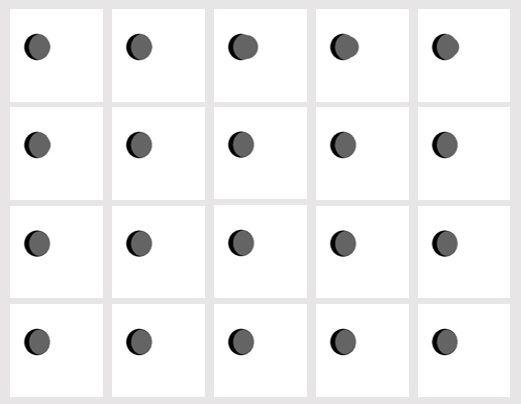}
\caption{Design \#1 on a synthetic image.}
\label{fig:synth1}
\end{figure}

\begin{figure}[H]
  \centering
  \subfloat[$I_0$]{\includegraphics[width=0.5\textwidth]{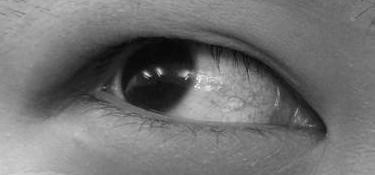}\label{fig:f1}}
  \hfill
  \subfloat[$I_1$]{\includegraphics[width=0.5\textwidth]{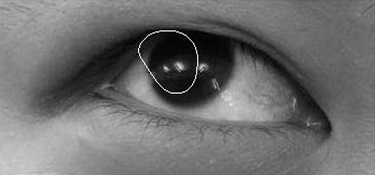}\label{fig:f2}}
  \caption{Design \#1}
  \label{fig:d1}
\end{figure}

Design \#1 minimizes the difference between the average intensity of the tracked region and the original region, and length to regularize the functional. This design was based on the supposition that the iris and pupil have an average intensity that is not matched by any nearby regions.
\begin{align*}
E_1=\lambda_1 \bigg[ \bigg( \frac{\iint_{R_1} I_1(x,y) \,dx\,dy}{\iint_{R_1} \,dx\,dy}\bigg)-\bigg(\frac{\iint_{R_0} I_0(x,y) \,dx\,dy}{\iint_{R_0} \,dx\,dy}\bigg)\bigg] ^2 + \lambda \int ||\dot{\gamma}(s)||  \,ds
\end{align*}
First this functional was tested on a synthetic image, shown in Figure \ref{fig:synth1}.
Design \#1 did not fully cover the synthetic region which is expected as any subset of the image with the same average intensity as the original region will minimize the intensity term in the functional.  Notice that the length term causes the tracked region to shrink as the number of iterations increases. Since Design \#1 behaved as expected when tested on a synthetic image, it was applied to a greyscale image of an eye seen in Figure \ref{fig:d1}.

In the test on the synthetic image, the tracked region only covered desired pixels, but in the test on a greyscale image, the functional failed to fully differentiate between shadowed regions and the eye. The accuracy metrics were:
\begin{center}
\textit{Desired Region Coverage} = 41\% \hspace{10mm} \textit{Undesired Region Coverage} = 10\%
\end{center}

\subsubsection{Design \#1B}
Design \#1B minimizes the difference between the average intensity and area of the tracked region and the original region, and length to regularize the functional. The second term in the functional was introduced to combat the problem in Design \#1 which caused the functional to be minimized by a subset of the desired region. 

\begin{align*}
E_{1B}=E_1+\lambda_2 \Big(\iint_{R_1} \,dx\,dy - \iint_{R_0} \,dx\,dy\Big)^2
\end{align*}

\begin{figure}[H]
\centering
\includegraphics[scale=0.8]{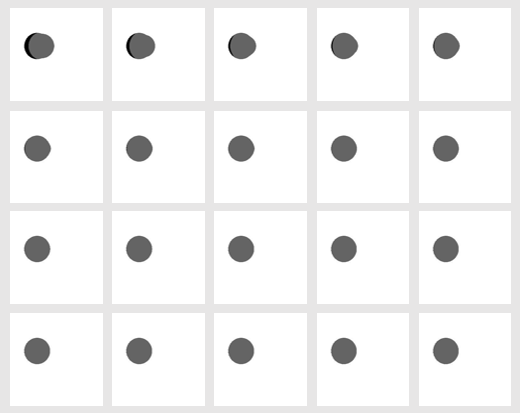}
\caption{Design \#1B on a synthetic image.}
\label{fig:synth1B}
\end{figure}

This functional was tested on a synthetic image, shown in Figure \ref{fig:synth1B}.

\begin{figure}[H]
  \centering
  \subfloat[$I_0$]{\includegraphics[width=0.5\textwidth]{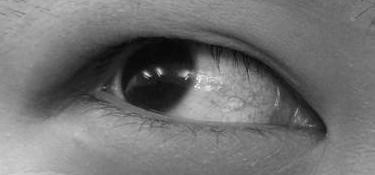}\label{fig:f1}}
  \hfill
  \subfloat[$I_1$]{\includegraphics[width=0.5\textwidth]{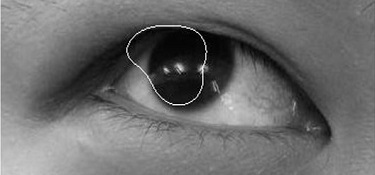}\label{fig:f2}}
  \caption{Design \#1B}
  \label{fig:d1b}
\end{figure}

Design \#1B performed very well on the uniform synthetic image, as it covered the desired region exactly. This was an improvement on the previous iteration. It was then applied to a greyscale image of an eye seen in Figure \ref{fig:d1b}.

When tested on an eye, this design had the same inability to differentiate between the desired region and other areas with similar intensity. This result is understandable as any combination of pixels which average to the desired value will minimize the functional. The accuracy metrics were:
\begin{center}
\textit{Desired Region Coverage}=56\% \hspace{10mm} \textit{Undesired Region Coverage}=24\%
\end{center}

\subsection{Design \#2}
Design \#2 is based on Kullback-Liebler divergence, and minimizes the “distance” between the intensity distribution of the tracked region and the original region, and the length to regularize the functional. This design was an attempt to combat the lack of information present in Design \#1 and \#1B. By minimizing the divergence the functional was looking at the intensity distribution which contains all the information about the intensities of the original region. 

\begin{align*}
E_2=\lambda_1 \int_\mathbb{I} p_{R_{\gamma_0}}(I) \log \frac{p_{R_{\gamma_0}}(I)}{p_{R_{\gamma_1}}(I)}+ \lambda \int ||\dot{\gamma}(s)||  \,ds
\end{align*}

\begin{figure}[H]
\centering
\includegraphics[scale=0.8]{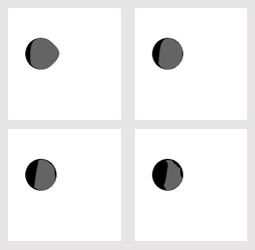}
\caption{Design \#2 on a synthetic image.}
\label{fig:synth2}
\end{figure}

This functional was tested on a synthetic image, shown in Figure \ref{fig:synth2}.

\begin{figure}[H]
  \centering
  \subfloat[$I_0$]{\includegraphics[width=0.5\textwidth]{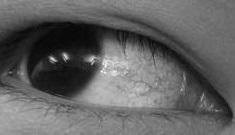}\label{fig:f1}}
  \hfill
  \subfloat[$I_1$]{\includegraphics[width=0.5\textwidth]{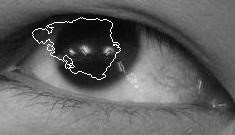}\label{fig:f2}}
  \caption{Design \#2}
  \label{fig:d2}
\end{figure}

Design \#2 had similar performance to Design \#1 on a uniform synthetic image which is understandable, as any subset of the desired region would give a similar distribution, and the length term would encourage smaller regions. The application of Design \#2 to a greyscale image of an eye seen in Figure\ref{fig:d2}.

On an eye it did not cover the entire region, and again failed to to fully differentiate between the area of heavier shadow and the desired region. The accuracy metrics were:
\begin{center}
\textit{Desired Region Coverage} = 55\% \hspace{10mm} \textit{Undesired Region Coverage} = 20\%
\end{center}

\subsubsection{Design \#2B}
Design \#2B is the same as Design \#2, plus a term to minimize the “distance” between the intensity distribution of the tracked region’s complement and the original region’s complement. This term was added in an attempt to both reduce the undesired region coverage and cover the entire eye. The complement term was intended to encourage pixels of non desired shaded regions to be excluded from the interior of the curve as they would be present in the original complement distribution. Similarly, desired regions might be encouraged into the interior of the curve due to them not being present in the initial complement distribution. 

\begin{align*}
E_{2B}=E_2+\lambda_2 \int_\mathbb{I} p_{R_{\gamma_0}^C} (I) \log \frac{p_{R_{\gamma_0}^C}(I)}{p_{R_{\gamma_1}^C}(I)}
\end{align*}

\begin{figure}[H]
\centering
\includegraphics[scale=0.8]{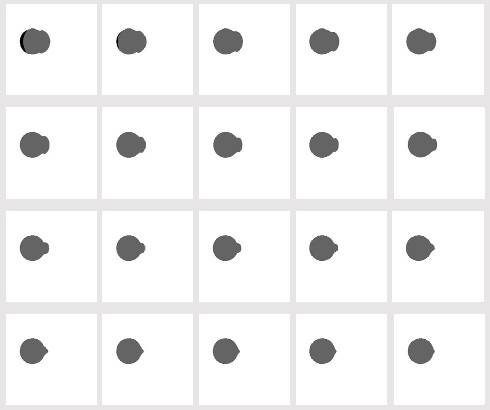}
\caption{Design \#2B on a synthetic image.}
\label{fig:synth2B}
\end{figure}

This functional was tested on a synthetic image, shown in Figure \ref{fig:synth2B}.

\begin{figure}[H]
  \centering
  \subfloat[$I_0$]{\includegraphics[width=0.5\textwidth]{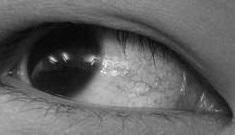}\label{fig:f1}}
  \hfill
  \subfloat[$I_1$]{\includegraphics[width=0.5\textwidth]{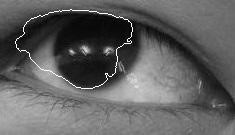}\label{fig:f2}}
  \caption{Design \#2B}
  \label{fig:d2b}
\end{figure}

On a synthetic image this functional performed well covering the desired region. Since Design \#2B behaved as expected when tested on a synthetic image, it was applied to a greyscale image of an eye seen in Figure\ref{fig:d2b}.

Design \#2B still could not differentiate between heavier shadowed regions and the actual eye. The functional displayed similar failings to Design \#1B where shadowed regions were covered undesirably. This phenomenon is understandable as any undesired pixel contained within the curve was essentially "swapped out" for pixels from the desired region with similar intensities. This allowed both the interior and exterior divergences to approach a minimum without covering the desired region. The accuracy metrics were:
\begin{center}
\textit{Desired Region Coverage} = 72\% \hspace{10mm} \textit{Undesired Region Coverage} = 29\%
\end{center}

\subsection{Design \#3}
Design \#3 again minimizes the “distance” between the intensity distribution of the tracked region and the original region, and the length to regularize the functional. It also minimizes the “distance” between the distribution of J of the tracked region and the original region, where J is the norm of the gradient of intensity given below. 

\begin{align*}
J=\bigg[\Big(\frac{\partial I}{\partial x}\Big)^2 +\Big(\frac{\partial I}{\partial y}\Big)^2\bigg]^{\frac{1}{2}}
\end{align*}

\begin{figure}[H]
\centering
\includegraphics[scale=0.8]{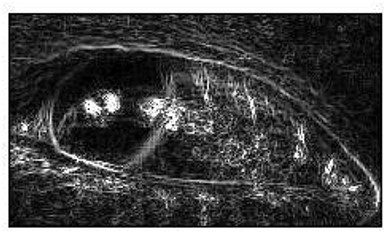}
\caption{Image of J values on $I_0$}
\label{fig:J}
\end{figure}
The new image produced by looking at J instead of intensity can be seen in Figure \ref{fig:J}. The functional was designed to use this edge information to more accurately find the region. In previous iterations, intensity information alone had not been successful. The design premise was that while a heavily shadowed region may have similar intensity to the desired portion of the eye, it would not have the same edges and might be more uniform. 

\begin{align*}
E_3=\lambda_1\int_\mathbb{I} p_{R_{\gamma_0}}(I) \log \frac{p_{R_{\gamma_0}}(I)}{p_{R_{\gamma_1}}(I)}+\lambda_2 \int_\mathbb{J} p_{R_{\gamma_0}}(J) \log \frac{p_{R_{\gamma_0}}(J)}{p_{R_{\gamma_1}}(J)}+ \lambda \int ||\dot{\gamma}(s)||  \,ds
\end{align*}

\begin{figure}[H]
\centering
\includegraphics[scale=0.8]{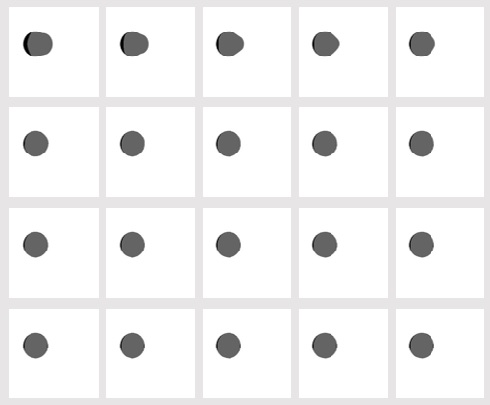}
\caption{Design \#3 on a synthetic image.}
\label{fig:synth3}
\end{figure}

This functional was tested on a synthetic image, shown in Figure \ref{fig:synth3}.

\begin{figure}[H]
  \centering
  \subfloat[$I_0$]{\includegraphics[width=0.5\textwidth]{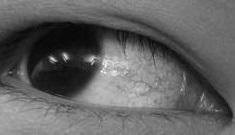}\label{fig:f1}}
  \hfill
  \subfloat[$I_1$]{\includegraphics[width=0.5\textwidth]{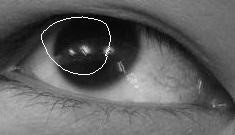}\label{fig:f2}}
  \caption{Design \#3}
  \label{fig:d3}
\end{figure}

Design \#3 mostly covered the desired region on a synthetic image, and was then applied to a greyscale image of an eye seen in Figure\ref{fig:d3}.

There was no motivation to cover the entire region but, more importantly, the functional again could not differentiate between shaded areas and the desired region. The accuracy metrics were: 
\begin{center}
\textit{Desired Region Coverage}=52\% \hspace{10mm} \textit{Undesired Region Coverage}=13\%
\end{center}

\subsection{Design \#4}
Design \#4 minimizes the “distance” between the RGB distribution of the tracked region and the original region, and the length to regularize the functional. This iteration of the design came from the realization that while the intensities of shaded areas and the desired region were similar, the colour was not. This realization lead to looking at the distribution of RGB colour intensities. This was done by adding up the divergence of each colour's distribution to form Design \#4. 

\begin{align*}
E_4=\lambda_1\int_\mathbb{I} p_{R_{\gamma_0}}(I_R) \log \frac{p_{R_{\gamma_0}}(I_R)}{p_{R_{\gamma_1}}(I_R)}+\lambda_2 \int_\mathbb{I} p_{R_{\gamma_0}}(I_G) \log \frac{p_{R_{\gamma_0}}(I_G)}{p_{R_{\gamma_1}}(I_G)}+\lambda_3\int_\mathbb{I} p_{R_{\gamma_0}}(I_B) \log \frac{p_{R_{\gamma_0}}(I_B)}{p_{R_{\gamma_1}}(I_B)}+\lambda \int ||\dot{\gamma}(s)||  \,ds
\end{align*}

\begin{figure}[H]
\centering
\includegraphics[scale=0.8]{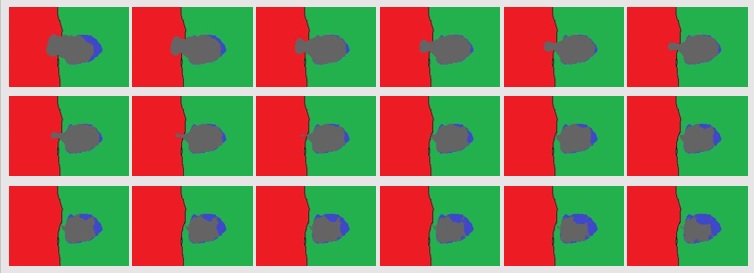}
\caption{Design \#4 on a synthetic image.}
\label{fig:synth4}
\end{figure}

This functional was tested on a synthetic image, shown in Figure \ref{fig:synth4}.

\begin{figure}[H]
  \centering
  \subfloat[$I_0$]{\includegraphics[width=0.5\textwidth]{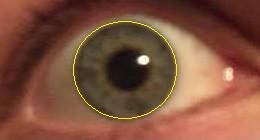}\label{fig:f1}}
  \hfill
  \subfloat[$I_1$]{\includegraphics[width=0.5\textwidth]{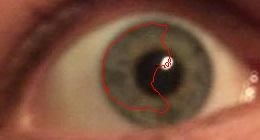}\label{fig:f2}}
  \caption{Design \#4}
  \label{fig:d4}
\end{figure}

On the synthetic image, this design showed that it could differentiate between different colours. Design \#4 was applied to a colour image of an eye seen in Figure\ref{fig:d4}.

Design \#4 added information that reduced local minima in the shadows, but did not cover the entire area. On an eye this design iteration performed significantly better than previous designs, with no undesired region coverage as seen below. However, once again there was no motivation to cover the entire desired region. 
\begin{center}
\textit{Desired Region Coverage}=43\% \hspace{10mm} \textit{Undesired Region Coverage}=0\%
\end{center}

\subsubsection{Design \#4B: Final Design}
Design \#4B again minimizes the “distance” between the RGB distribution of the tracked region and the original region, and the length to regularize the functional. It also includes a term to minimize the difference between the area of the tracked region and the original region. 
\begin{align*}
E_{4B}=E_4+\lambda_4 \Big(\iint_{R_1} \,dx\,dy - \iint_{R_0} \,dx\,dy\Big)^2
\end{align*}

\begin{figure}[H]
\centering
\includegraphics[scale=0.8]{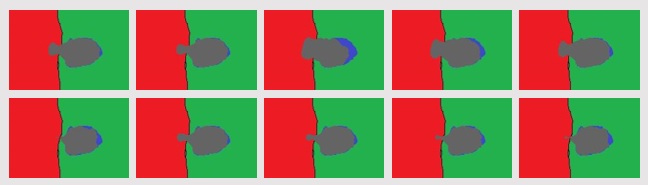}
\caption{Design \#4B on a synthetic image.}
\label{fig:synth4B}
\end{figure}

This functional was tested on a synthetic image, shown in Figure \ref{fig:synth4B}.

\begin{figure}[H]
\centering
\includegraphics[scale=0.8]{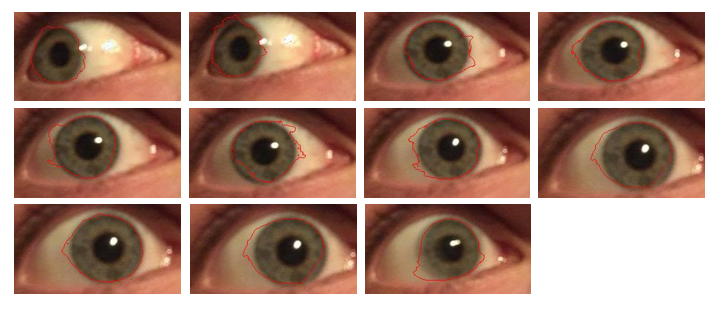}
\caption{Design \#4B Results Set 1.}
\label{fig:result1}
\end{figure}

\begin{figure}[H]
\centering
\includegraphics[scale=0.8]{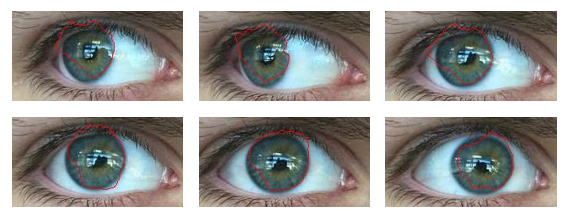}
\caption{Design \#4B Results Set 2.}
\label{fig:result2}
\end{figure}

Design \#4B covered covered most of the desired region in synthetic testing but had a small amount of undesired region coverage. It was applied to multiple colour images of an eye seen in Figures \ref{fig:result1} and \ref{fig:result2}.

Design \#4B gives an approximation of the location of the desired region through a sequence of images. There is a trade-off present in tuning the parameters of this design between region coverage and undesired region coverage. Higher $\lambda_4$ relative to the other multipliers resulted in more desired region coverage, but also resulted in more undesired region coverage. This final design also requires customization of all multipliers in order to function of different eyes. The average accuracy metrics for the above results were:
\begin{center}
\textit{Desired Region Coverage} = 82\% \hspace{10mm} \textit{Undesired Region Coverage} = 20\%
\end{center}
The final design is able to provide approximate eye location data to a larger system, and clearly gave better results than all of the preceding designs.

\section{Next Steps}
In order to incorporate the final design in an effective driver alertness system, the design needs further work. The algorithm needs to be able to run in real time in order to track the driver's eye. This could be partially accomplished through parallel computing, as for each time step $\delta$t of the algorithm, the evaluation of the functional at each pixel is independent of the surrounding pixels. The function will also need further refinement in order to give an accurate approximation of the drivers gaze. This is necessary to prevent the system from warning and annoying an attentive driver, as well as to make sure it catches all distraction. The algorithm currently requires tooling to each individual eye that it operates on. This is not acceptable for an implementable solution as it must work on all drivers. As such, the functional needs to be modified and tested on a wider range of eye colours and shapes to make sure it operates reliably. Finally, the algorithm developed in this project needs to be integrated into a larger system that takes the information provided by tracking the desired region and determines the direction of gaze. From this direction of gaze the system must further determine if such behavior is acceptable or represents a distraction before warning the driver.

\section{Driver Alertness} 
\subsection{Environmental Impact}
%Impact of engineering/Consider social and environmental factors (Rubric Section): Shows superior understanding of economic, social, and environmental factors and/or impacts in decisions on specific application area.
The system will require power, leading to increased carbon emissions. The power use will depend on how accurate and fast the algorithm is required to be, combined with the efficiency of the solution. This could be mitigated by the use of electric cars and renewable energy, but currently would result in increased fossil fuel consumption. \\
\\The design can reduce negative impacts on the environment due to vehicle accidents. Traffic accidents cause congestion, spills of gas and other harmful substances, and wreck cars. Approximately 15\% of congestion is due to traffic accidents\cite{conjest2010} and congestion adds on average 7\% to greenhouse gas emissions from driving \cite{ghgconjest2008}.This does not take into account the carbon emissions from the emergency repsonse vehicles.
\\%One in four accidents are caused by texting and driving alone, this does not take into account other forms of distracted driving. Every time there is an accident, it causes a lot of cars to move over, and a lot of gasoline is burned in the process. Not just from the vehicles idling, but also from the emergency vehicles that need to rush over.
In the United States, 1.6 million crashes are due to cell phone use alone \cite{edgar}. On average it takes approximately 1.3 GJ to repair a car \cite{fridley}, so this totals to 2,176,000 GJ per year. Which is equivalent to operating 650,000 homes for a month in the United States\cite{usenergy}.\\ 
%\\The eye tracking device can also alert sleep deprived/drowsy drivers. The National Highway Traffic Safety Administration estimates that 100,000 police-reported crashes in the USA are the direct result of driver fatigue each year. This results in approximately 1,550 deaths, 71,000 injuries, and 12.5 billion in monetary losses. These figures may be a lot lower than what it actually is, since currently it is difficult to attribute crashes to sleepiness\cite{drowdriv}.

\subsection{Economic Impact}
%Economics \& Project Management/Performs Economic Analysis (Rubric Section): Performs an economic analysis on multiple solutions and uses quantitative justification when choosing solution.
78\% of traffic accidents are related to driver inattention \cite{prevent2011} and at least 25\% are caused by it \cite{sensitive2005}. This indicates that a successful implementation can have a sizable impact on cost related to collisions.\\
\\In 2010, vehicle accidents caused the USA lost worker productivity costs of \$185, property damage cost of \$245, and congestion cost of \$90, per capita in USD \cite{colcost2010}. In Canada, the total combined cost of collisions is approximately 2\% of GDP\cite{colcan2011} (US \$31 Billion\cite{cangdp2013}). This indicates that Canadians could save up to 7.75 Billion dollars, based on 25\%  of accidents being caused by driver inattention.\\
%At the assumed rate of success, the system could save \$18.5, \$24.5. In total, this amounts to \$1.83 billion in savings, which is about 7.5\% of the total cost of collisions.
\\According to the National Highway Traffic Safety Administration, there were approximately 100,000 crashes in the USA due to driver fatigue each year. Of this there were roughly 1550 deaths, and 71,000 injuries which lead to over US\$12.5 billion in monetary losses\cite{drowdriv}.\\
\\Drowsy driving also occurs in work place environments as well. For instance, in the commercial trucking industry, driver inattention caused 1,200 deaths and 76,000 injuries each year, which is equivalent to a total cost of US\$10.3 billion. Because 65\% of all these accidents are fatigue-related, the designed system can potentially create a savings of US\$6.7 billion\cite{comtruck}.

\subsection{Societal Impact}
%Impact of engineering/Consider social and environmental factors (Rubric Section): Shows superior understanding of economic, social, and environmental factors and/or impacts in decisions on specific application area.

\subsubsection{Safety Considerations}
In 2014 there were 1,834 vehicle accident fatalities and 9,647 serious injuries in Canada \cite{canmot2014}. In terms of years of life lost compared to the average lifespan, vehicle fatalities accounts for 3.1\% of Canada’s total years of life lost \cite{gdbprof2012}. This shows that a reduction in distracted driving could potentially have a major impact on length of life in Canada. A reliable and effective design is imperative because not only is the user’s life at stake, the system also impacts the other vehicle occupants on the road.\\
\\Vehicle collisions are a major burden on the health care system in Canada \cite{colcan2011}. In 2009 direct health care costs from vehicle collisions accounted for CA \$10.7 billion \cite{ecburd2009}. This is .52\% of the total health care expenditure at that time \cite{healthcost2015}. Reducing collisions could therefore free up resources for other medical programs.
\subsubsection{People at Risk}
Citizens in all levels of society are affected by motor vehicle crashes - the individual crash victims and their families, their employers, and society at large. Victims suffer from physical pain, disability, and emotional impacts that can greatly reduce the quality of their lives. It also negatively impacts the victim’s dependents due to an immediate economic hardship in the loss of the victim’s income and other contributions. In total, the value of societal harm, from motor vehicle crashes for lost quality-of-life, in the United States, was US\$594 billion in 2010 \cite{NHTSA}.
\subsubsection{Privacy Concerns}
The proposed design will have privacy implications due to its constant tracking of driver’s eyes. In the USA law enforcement officials already use data such as text history to investigate distracted driving after a collision \cite{txtdrive2013}. If the new system is implemented, data on the attentiveness of drivers may be demanded, which would help determine fault in collision, but could compromise people’s personal privacy.

\subsection{Ethics and Equity}
%Ethics and Equity/Considers ethical factors (Rubric Section): Shows superior understanding and evaluates ethical factors and matters of equity concepts. 

The greatest ethical concerns are the lives that are at risk due to inattentive and drowsy drivers. As stated before, 1,834 deaths occurred in Canada in 2014 \cite{canmot2014}. With 10\% accuracy, the system could save up to 183 lives per year. This is a huge improvement and should be weighted very heavily when considering the ethical implications. Not only are the drivers who are at fault for dangerous driving at risk, but other drivers and people or children walking on sidewalks are also at great risk of injury or death in these accidents. Approximately 20\% (366) of the 1,834 deaths were pedestrian or bicyclists \cite{canmot2014}.\\
\\Many agencies, companies, or individuals with malicious intent, can hack into almost any programed system. It has been shown that it is possible to access a car's computer and control its brakes, a/c, lights, etc \cite{hacker2015}. Many news releases have suggested proof of the NSA and CIA infecting systems with programs that allow them to track the users' camera, voice recorder, or any various sensors on the machine. It is sensible to then ask about the ability for external sources to track and gain live or past information using illegal methods through the eye tracking system. There are currently multiple court cases going on in the USA and Canada which could potential allow these companies to gain access to this information using these various methods. \cite{trump2017}. As mentioned before, ethical issues are also raised when external sources can gain access to the systems information without consent (police obtain a warrant). This system opens another opportunity for authorities to acquire personal information from drivers regardless of the drivers wants. \\
\\Questions of driver equity also must be raised when implementing the system. Many engineers and staticians believe that autonomous vehicles will be mandatory in the next few decades, while manual driving will be illegal, as autonomous driving has been proven to reduce the risks of driving \cite{human2016}. Laws have been developed for seatbelt alerts, making it mandatory for the car to alert the driver/ passengers when their seatbelt is not clipped in \cite{seatbelt2017}. Similar laws could be made for the eye tracking system, depending on its success. This would involve every car having a mandatory video monitoring system installed in it. The other two viable options are having the system optional for all drivers, or having the system mandatory for only a subset of drivers \cite{dui2014}. This could include drivers with a accident record (similar to breathalyzers installed on the cars owned by people with DUI’s), or based off of their age, driving experience, car model, etc. These decisions further open up the floor for biased decision making which can reduce the equity for drivers on the road. It is unethical to force certain people to use this system unless it is proven that they produce a greater danger on the road than other drivers.

\subsection{Design Trade-Offs}
%Impact of engineering/Evaluates trade-offs (Rubric Section): Shows superior and/or novel means of evaluating trade-offs among goals and concepts.
 
The proposed design objective will alert the driver when it notices unacceptable eye positions. This design must be able to quickly identify eye direction that indicate distraction, as on a highway cars can travel over 30 m in a second. To achieve this a functional was designed for computational efficiency at the forefront of the design process.\\

Functionals with higher order derivatives than first order were ignored, as this would greatly increase the computational time while only increasing the eye location accuracy by a few percent. Minimization of the distribution of curvature was avoided for the same reason. \\

	Another trade-off of the functional is the dependence on the RGB distribution. As can be seen between the final design and the preceding ones, the RGB distribution term is essential to the accuracy of the desired region tracking. The issue lies with using the functional for various colored eyes. As mention before, one option is to conduct tests on various eye colors and have the system able to learn which multipliers ($\lambda_1$, $\lambda_2$, $\lambda_3$) to apply for each eye color. Thus, the eye tracking program could pick the associated multipliers  when the driver starts the car and the camera picks up the eye color of the driver, before they start moving. \\

	The computation time was decreased by using a linear combination of the RGB distributions instead of the full distribution itself. By adding the red distribution, green distribution, and blue distribution, with scalar multipliers, the computation was significantly quicker than using a 3x3 distribution of the full RGB distribution. The issue here is the individual scalar multipliers require fine tuning all three multipliers individually, instead of one multiplier for the whole distribution. While the computation time was decreased, it has become more difficult to use the program for all eye colors, as mentioned before.

\subsection{Regulatory Concerns}
%Professionalism/Integrates Standards (Rubric Section): Shows a superior knowledge and application of standards, codes of practice, and legal and regulatory factors in the decision-making processes (as appropriate).
In Ontario, it is against the law to operate hand-held communication or electronic entertainment devices while driving\cite{mintrans}. Simply holding a phone or other device while driving is against the law. Other activities considered to be distracted driving include eating and reading\cite{distdriv}. If convicted, the penalty for distracted driving depends on the type of licence possessed by the driver. A fully licenced driver (holder of Class A, B, C, D, E, F, G licence) will receive a fine of \$400, plus a victim surcharge and court fee, for a total of \$490 if settled out of court. There will be a fine of up to \$1,000 if the driver receives a summons or fight their ticket. Lastly, three demerit points will be applied to their driver’s record. If a novice driver (holder of G1, G2, M1 or M2 licence) is convicted of distracted driving, they will receive the same fines as drivers with A to G licences. But will not receive any demerit points. Instead they will be faced with a 30-day licence suspension for a first conviction, a 90-day licence suspension for a second conviction, and a cancellation of their licence and removal from the Graduated Licensing System for a third conviction.\\ 
\\Drivers who endanger others due to any kind of distraction (this includes the use of hands-free devices as well as hand-held devices), may face more charges for Careless Driving under the Highway Traffic Act or even Dangerous Driving under the Criminal Code of Canada. If convicted of Careless Driving, drivers may receive six demerit points, fines up to \$2000 and/or a jail term of six months, and up to a two-year licence suspension. Dangerous Driving is a criminal offence and includes jail terms of up to 10 years for causing bodily harm or up to 14 years for causing death\cite{distdriv}.\\
\\
Taking these laws into consideration, the physical system should be designed so that it is completely hands-free and is activated as the engine is turned on. The only action the user must take will be to adjust the camera. It will take on a similar function to the seatbelt alerting mechanism where a buzzer will go off when the driver is not paying attention to the road.

\subsection{Potential Impact of System}
If the system is implemented to all vehicles and assuming it has a conservative success rate of 10\% at alerting drivers that are distracted, it can mitigate negative impacts. Using statistics from 2014, in Canada, 183 lives could be saved, and 950 injuries\cite{canmot2014} prevented per year. Furthermore, there could be savings of CA\$125 million per year in monetary losses for vehicle collisions \cite{drowdriv}. Lastly, since motor vehicle travel accounts for 20\% of GHG emissions in Canada\cite{ghgcan2014}, a reduction of up to 0.02\% of Canada’s emissions will be possible.

\section{Conclusion}
In this project an algorithm was created that can give an approximation of the location of a subject’s iris and pupil. On average the final functional resulted in 82\% coverage of the desired region with 20\% of the tracked region not overlapping the iris and pupil. This design requires more work to become an effective component of a driver inattention preventing system. When implemented such a system has the potential to have significant positive environmental, economic, and social impacts. 

\newpage
\begin{flushleft}
\bibliographystyle{ieeetr}
\bibliography{citations}
\end{flushleft}

\end{document}